\pdfoutput=1

\documentclass[11pt]{article}

\usepackage[]{acl}

\usepackage{times}
\usepackage{latexsym}

\usepackage[T1]{fontenc}

\usepackage[utf8]{inputenc}

\usepackage{microtype}
\usepackage{amsmath}
\usepackage{graphicx}
\usepackage{multirow}
\usepackage{amsfonts}
\title{Field Extraction from Forms with Unlabeled Data}

\author{\textbf{Mingfei Gao}, \textbf{Zeyuan Chen}, \textbf{Nikhil Naik}, \textbf{Kazuma Hashimoto}, \textbf{Caiming Xiong}, \textbf{Ran Xu} \\
Salesforce Research, Palo Alto, USA \\
{\tt\small{\{mingfei.gao, zeyuan.chen, nnaik, k.hashimoto, cxiong, ran.xu\}@salesforce.com}}
  }

\begin{document}
\maketitle
\begin{abstract}
We propose a novel framework to conduct field extraction from forms with unlabeled data. To bootstrap the training process, we develop a rule-based method for mining noisy pseudo-labels from unlabeled forms. Using the supervisory signal from the pseudo-labels, we extract a discriminative token representation from a transformer-based model by modeling the interaction between text in the form. To prevent the model from overfitting to label noise, we introduce a refinement module based on a progressive pseudo-label ensemble. Experimental results demonstrate the effectiveness of our framework.
\end{abstract}

\section{Introduction}
\begin{figure}
    \centering
    \includegraphics[width=1.0\linewidth]{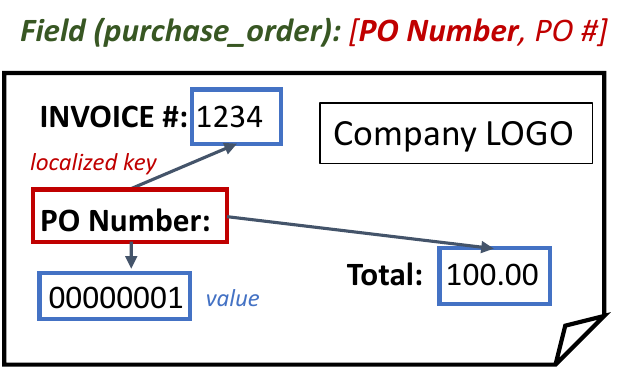}
    \caption{Field extraction from forms is to extract the value for each field, e.g., \emph{invoice$\_$number}, \emph{purchase$\_$order} and \emph{total$\_$amount}, in a given list. A key, e.g., INVOICE$\#$, PO Number and Total, refers to a concrete text representation of a field in a form and it is an important indicator for value localization. 
    }
    \label{fig:illustration}
\end{figure}

Form-like documents, such as invoices, paystubs and patient referral forms, are very common in daily business workflows. A large amount of human effort is required to extract information from forms every day. In form processing, a worker is usually given a list of expected form fields (e.g., \emph{purchase\_order}, \emph{invoice\_number} and \emph{total\_amount} in Figure~\ref{fig:illustration}), and the goal is to extract their corresponding values based on the understanding of the form, where keys are generally the most important features for value localization. A field extraction system aims to automatically extract field values from redundant information in forms, which is crucial for improving processing efficiency and reducing human labor.

Field extraction from forms is a challenging task. Document layouts and text representations can be very different even for the same form type, if they are from different vendors. For example, invoices from different companies may have significantly different designs (see Figure~\ref{fig:vis}). Paystubs from different systems (e.g., ADP and Workday) have different representations for similar information.

Recent methods formulate this problem as field-value pairing or field tagging. \citet{majumder2020representation} propose a representation learning method that takes field and value candidates as inputs and utilizes metric learning techniques to enforce high pairing score for positive field-value pairs and low score for negative ones. LayoutLM~\cite{xu2020layoutlm} is a pretrained transformer that takes both text and their locations as inputs. It can be used as a field-tagger which predicts field tags for input texts. These methods show promising results, but they require large amount of field-level annotations for training. Acquiring field-level annotations of forms is challenging and sometimes even impossible since (1) forms usually contain sensitive information, so there is limited public data available; (2) working with external annotators is also infeasible, due to the risk of exposing private information and (3) annotating field-level labels is time-consuming and hard to scale. 

Motivated by these reasons, we propose a field extraction system that does not require field-level annotations for training (see Figure~\ref{fig:our_method}). First, we bootstrap the training process by mining pseudo-labels from unlabeled forms using simple rules. Then, a transformer-based architecture is used to model interactions between text tokens in the form and predict a field tag for each token accordingly. The pseudo-labels are used to supervise the transformer training. Since the pseudo-labels are noisy, we propose a refinement module to improve the learning process. Specifically, the refinement module contains a sequence of branches, each of which conducts field tagging and generates refined labels. At each stage, a branch is optimized by the labels ensembled from all previous branches to reduce label noise. Our method shows strong performance on real invoice datasets. Each designed module is validated via comprehensive ablation experiments.

Our contribution is summarized as follows: (1) to the best of our knowledge, this is the first work that addresses the problem of field extraction from forms without using field-level labels; (2) we propose a novel training framework where simple rules are first used to bootstrap the training process and a transformer-based model is used to improve performance; (3) our proposed refinement module is demonstrated as effective to improve model performance when trained with noisy labels and (4) to facilitate future research, we introduce the INV-CDIP dataset as a public benchmark. The dataset is available at \href{https://github.com/salesforce/inv-cdip}{https://github.com/salesforce/inv-cdip}.

\section{Related Work}
\subsection{Form understanding} 
Form understanding is a widely researched area. Earlier work formulated the problem as an instance segmentation task. Chargrid~\cite{katti2018chargrid} encodes each page of form as a two-dimensional grid of characters, and extracts header and line items from forms using fully convolutional networks. Based on Chargrid,~\citet{denk2019bertgrid} propose BERTgrid which uses a grid of contextualized word embedding vectors to represent documents. These methods are limited in scenarios where the image resolution is not high enough leading to sub-optimal representation of ambiguous structures in dense regions. To mitigate the issue, later methods work on structure modeling. \citet{aggarwal2020form2seq} introduce Form2Seq to leverage relative spatial arrangement of structures via first conducting low-level element classification and then high-order grouping. DocStruct~\cite{wang2020docstruct} encodes the form structure as a graph-like hierarchy of text fragments and designs a hybrid fusion method to provide joint representation from multiple modalities. Benefiting from the recent advances of transformers~\cite{vaswani2017attention}, LayoutLM~\cite{xu2020layoutlm} learns text representation via modeling the interaction between text tokens and their locations in documents. 

There are dedicated methods focusing on field extraction. Some methods~\cite{chiticariu-etal-2013-rule, 6628593} extract information from document via registering templates in the system. \citet{palm2019attend} propose an Attend, Copy, Parse architecture to extract field values of invoices. ~\citet{majumder2020representation} present a metric learning framework that learns the representation of the value candidate based on its nearby words and matches the field-value pairs using a learned scoring function. \citet{gao2021value} propose a general value extraction system for arbitrary queries and introduce a simple pretraining strategy to improve document understanding.
Although existing approaches demonstrate promising results in different settings, they rely on large-scale annotated data for training. For example,~\citet{majumder2020representation} used more than 11,000 invoices in distinct templates for training.

\subsection{Form datasets} 
Form datasets for field extraction tasks are typically private, since these documents generally contain sensitive information. There are existing public datasets for general form understanding. RVL-CDIP~\cite{harley2015icdar} and DocVQA~\cite{mathew2021docvqa} are introduced for document classification and question answering tasks. FUNSD~\cite{jaume2019funsd} dataset is organized as a list of interlinked semantic entities, i.e., question, answer, header and other. CORD~\cite{park2019cord} is a public receipt dataset focusing on line items. SROIE~\cite{huang2019icdar2019} is the most related dataset which aims to extract information for four receipt-related fields. However, their layouts across different receipts are very fixed, which makes it less challenging, thus not suitable for our task. For example, the values of fields, \emph{company} and \emph{address}, are always on the very top in all the receipts. The lack of appropriate public datasets makes it difficult to compare existing field extraction methods on realistic forms. \citet{xue2021robustness} introduce a framework to augment diverse forms from a small set of annotated forms for robust evaluation. In this work, we introduce a challenging and real invoice dataset that is made publicly available to future research.

\begin{figure*}
    \centering
    \includegraphics[width=1.0\linewidth]{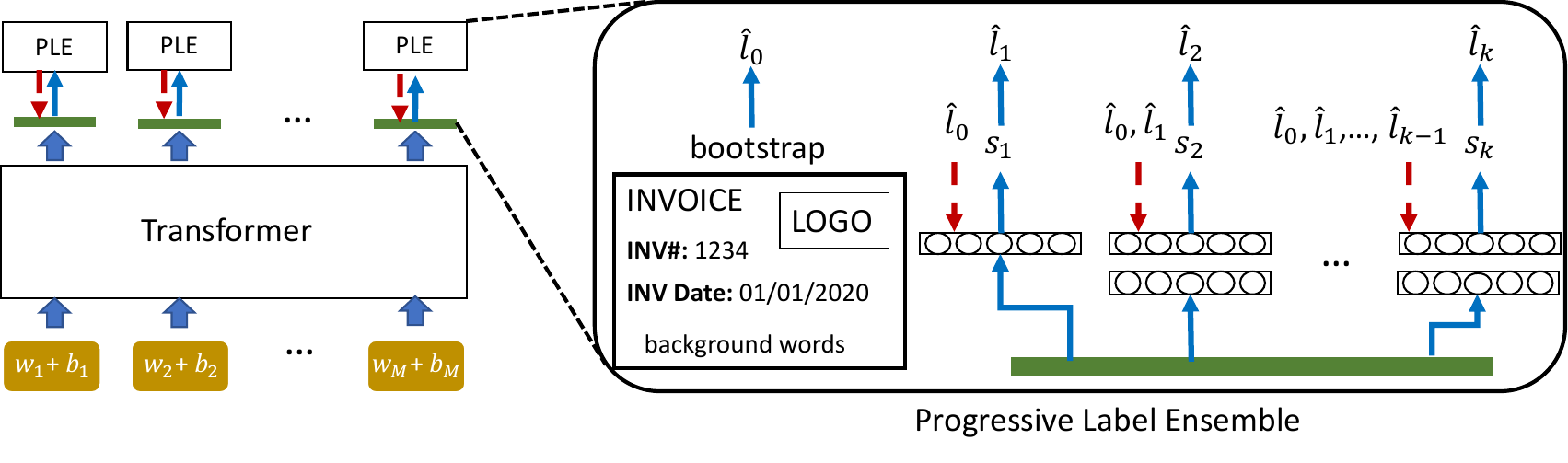}
    \caption{Our method takes words, $w_i$, and their locations, $b_i$, in a form into a transformer. The transformer extracts representative features for each token via the self-attention mechanism. Since our method is trained using forms with no field labels, we design progressive label ensemble module to enable the training process. We bootstrap the initial pseudo-labels, $\hat{l}_0$, using simple rules. Then, token representations go through several branches and do the field prediction as well as label refinement. Each branch, $j$, is optimized with labels ensembled from all previous branches, $\hat{l}_0, \hat{l}_1,...,\hat{l}_{j-1}$.}
    \label{fig:our_method}
\end{figure*}

\section{Field Extraction from Forms}
\subsection{Problem Formulation}
We are interested in information of fields in a predefined list, $\{fd_1, fd_2,...,fd_N\}$. Given a form as input, a general OCR detection and recognition module is applied to obtain a set of words, $\{w_1, w_2,...,w_M\}$, with their locations represented as bounding boxes, $\{b_1, b_2,...,b_M\}$.
The goal of a field extraction method is to automatically extract the target value, $v_i$, of field, $fd_i$, from the massive word candidates if the information of the field exists in the input form.

Unlike previous methods that have access to large-scale labeled forms, the proposed method can be trained using unlabeled documents with known form types. To achieve this goal, we propose a simple rule-based method to mine noisy pseudo-labels from unlabeled data (Sec.~\ref{sec:pseudo_labels}) and introduce a data-driven method with a refinement module to improve training with noisy labels (Sec.~\ref{sec:refinement}).

\subsection{Bootstrap: Pseudo-Labels Inference from Unlabeled Data}
\label{sec:pseudo_labels}
To bootstrap the training process, given unlabeled forms, we first mine pseudo-labels using a simple rule-based algorithm. The algorithm is motivated by the following observations: (1) a field value usually shows together with some key and the key is a concrete text representation of the field (see Figure~\ref{fig:illustration}); (2) the keys and their corresponding values have strong geometric relations. As shown in Figure~\ref{fig:illustration}, the keys are mostly next to their values vertically or horizontally; (3) although the form's layout is very diverse, there are usually some key-texts that frequently used in different form instances. For example, the key-texts of the field \emph{purchase\_order} can be ``PO Number", ``PO \#" etc. and (4) inspired by~\citet{majumder2020representation}, the field values are always associated with some data type. For example, the data type of values of "invoice\_date" is \emph{date} and that of "total\_amount" is \emph{money} or \emph{number}. 

Based on the above observations, we design a simple rule-based method that can efficiently get useful pseudo-labels for each field of interest from large-scale forms. As shown in Figure~\ref{fig:illustration}, key localization is first conducted based on string-matching between text in a form and possible key strings of a field. Then, values are estimated based on data types of the text and their geometric relationship with the localized key.

\noindent\textbf{Key Localization}. Since keys and values may contain multiple words, we obtain phrase candidates, $[ph_i^1, ph_i^2,...,ph_i^{T}]$, and their locations $[B_i^1, B_i^2,...,B_i^{T}]$ in the form by grouping nearby recognized words based on their locations using DBSCAN algorithm~\cite{ester1996density}. For each field of interest, $fd_i$, we design a list of frequently used keys, $[k_i^1, k_i^2, ..., k_i^L]$, based on domain knowledge. In practice, we can also use the field name as the only key in the list. Then, we measure the string distance\footnote{Without loss of generality, Jaro–Winkler distance~\cite{winkler1990string} is used in this work.} between a phrase candidate, $ph_i^{j}$, and each designed key, $k_i^r$, as $d(ph_i^{j}, k_i^r)$. We calculate the key score for each phrase candidate indicating how likely this candidate is to be a key for the field using Eq.~\ref{eq:keyscore}. Finally, the key is localized by finding the candidate with the largest key score as in Eq.~\ref{eq:findkey}.

\begin{equation}
\label{eq:keyscore}
    key\_score(ph_i^{j}) = 1 - \underset{r\in\{1,2,...,L\}}{min}d(ph_i^{j}, k_i^r).
\end{equation}

\begin{equation}
\label{eq:findkey}
    \hat{k}_i = \underset{j\in\{1,2,...,T\}}{argmax}key\_score(ph_i^{j}).
\end{equation}
\noindent\textbf{Value Estimation}. Values are estimated following two criteria. First, their data type should be in line with their fields. Second, their locations should accord well with the localized keys. For each field, we design a list of eligible data type (see Table~\ref{tab:keylist} in the appendix, Sec.~\ref{sec:appendix}). A pretrained BERT-based NER model~\cite{devlin2018bert} is used to predict the data type of each phrase candidate and we only keep the candidates, $ph_i^j$, with the correct data type.

Next, we assign a value score for each eligible candidate, $ph_i^j$ as in Eq.~\ref{eq:valuescore}, where $key\_score(\hat{k}_i)$ indicates the key score of the localized key and $g(ph_i^j, \hat{k}_i)$ denotes the geometric relation score between the candidate and the localized key. Intuitively, the key and its value are generally close to each other and the values are likely to just beneath the key or reside on their right side as shown in Figure~\ref{fig:illustration}. So, we use distance and angles to measure key-value relation as shown in Eq.~\ref{eq:keyvaluegeo}, where $dist_i^{j\rightarrow r}$ indicates the distance of two phrases, $angle_i^{j\rightarrow r}$ indicates the angle from  $ph_i^j$ to $ph_i^r$ and $\Phi(.|\mu,\,\sigma)$ indicates Gaussian function with $\mu$ as mean and $\sigma$ as standard deviation. Here, we set $\mu_d$ to 0. $\sigma_d$ and $\sigma_a$ are fixed to 0.5. We want to reward the candidates whose angle with respect to the key is close either to 0 or $\pi/2$, so we take the maximum angle score of these two options.

\begin{equation}
\label{eq:valuescore}
    value\_score(ph_i^{j}) = key\_score(\hat{k}_i)*g(\hat{k}_i, ph_i^j).
\end{equation}

\begin{equation}
\label{eq:keyvaluegeo}
\begin{split}
    g(ph_i^j, ph_i^r) &=\Phi(dist_i^{j\rightarrow r}|\mu_d,\,\sigma_d) \\
    &+\alpha\underset{\mu_a \in \{0, \pi/2\}}{max}\Phi(angle_i^{j\rightarrow r}|\mu_a,\,\sigma_a).
\end{split}
\end{equation}

\begin{equation}
\label{eq:findvalue}
    \hat{v}_i = \underset{j\in\{1,2,...,T\}, ph_i^j\neq\hat{k}_i}{argmax}value\_score(ph_i^{j}).
\end{equation}
We determine a candidate as the predicted value for a field if its value score is the largest among all candidates as in Eq.~\ref{eq:findvalue} and the score exceeds a threshold, $\theta_v=0.1$.

\subsection{Refinement with Progressive Pseudo-Labels Ensemble (PLE)}
\label{sec:refinement}
The above rule can be used directly as a simple field extraction method. To further improve performance, we can learn a data-driven model using the estimated values of fields as pseudo-labels during training. We formulate this as a token classification task, where the input is a set of tokens extracted from a form and the output is the predicted field including background for each token.

\noindent\textbf{Feature Backbone}.
To predict the target label of a word, we need to understand the meaning of this word as well as its interaction with the surrounding context. Transformer-based architecture is a good fit to learn the word's representation for its great capability of modeling contextual information. Except for the semantic representation, the word's location and the general layout of the input form are also important and could be used to capture discriminative features of words. In practice, we used the recently proposed LayoutLM~\cite{xu2020layoutlm} as the default backbone and also experimented with other transformer-based structures in Sec.~\ref{sec:experiments}.

\noindent\textbf{Field Classification}.
Field prediction scores, $s_k$, are obtained by projecting the features to the field space ($\{background, fd_1, fd_2,...,fd_N\}$) via fully connected (FC) layers.

\noindent\textbf{Progressive Pseudo-Labels Ensemble}.
Initial word-level field labels (also referred to as Bootstrap Labels), $\hat{l}_0$, are obtained by the estimated pseudo-labels from Sec.~\ref{sec:pseudo_labels} and the network can be optimized using cross entropy loss, $L(s_k, \hat{l}_0)$. However, naively using the noisy labels can degrade the model performance. We introduce a refinement module to tackle this issue. As shown in Figure~\ref{fig:our_method}, we use a sequence of classification branches, where each branch, $j$, conducts field classification independently and refines pseudo-labels, $\hat{l}_j$, based on their predictions. A later-stage branch is optimized using the refined labels obtained from previous branches. The final loss aggregates all the losses as
\begin{equation}
\label{eq:loss}
    L_{total} = L(s_1, \hat{l}_0) + \displaystyle\sum_{k=2}^{K}\displaystyle\sum_{j=1}^{k-1}(L(s_k, \hat{l}_j)+\beta L(s_k, \hat{l}_0)),
\end{equation}
where $\beta$ is a hyper parameter controlling the contribution of the initial pseudo-labels.

At branch $k$, we generate refined labels according to the following steps: (1) find the predicted field label, $\hat{fd}$, for each word by $\underset{c\in\{0,1,...,N\}}{argmax}s_{kc}$ and (2) for each positive field, only keep the word if its prediction score is the highest among all the words and larger than a threshold (fixed to 0.1).

Intuitively, each branch can be improved by using more accurate labels and its generated labels are further refined. This progressive refinement of labels reduces label noise. Similar idea has been used in weakly supervised object detection~\cite{tang2017multiple}. However, we find that using only the refined labels in each stage is limited in our setting, because although the labels become more precise after refinement, some low-confident values are filtered out which results in lower recall. To alleviate this issue, we optimize a branch with the ensembled labels from all previous stages. We believe that the ensembeled labels can not only keep a better balance between precision and recall, but also are more diverse and can serve as a regularization for model optimization. During inference, we use the average score predicted from all branches. We follow the same procedure to get final field values as we generate refine labels.

\section{Experiments and Results}
\label{sec:experiments}
\subsection{Datasets}
\noindent\textbf{IN-Invoice Dataset}.
We internally collect real invoices from different vendors. These invoice images are converted from real PDFs, so they are in high resolution with clean background. The train set contains 7,664 \textbf{unlabeled} invoice forms of 2,711 vendors. The validation set contains 348 \textbf{labeled} invoices of 222 vendors. The test set contains 339 \textbf{labeled} invoices of 222 vendors. We manually ensure that at most 5 images are from the same vendor in each set. Following~\citet{majumder2020representation}, we consider 7 frequently used fields including \emph{invoice\_number}, \emph{purchase\_order}, \emph{invoice\_date}, \emph{due\_date}, \emph{amount\_due}, \emph{total\_amount} and \emph{total\_tax}.

\noindent\textbf{INV-CDIP}.
This dataset is from the Tobacco Collections of Industry Documents Library~\footnote{https://www.industrydocuments.ucsf.edu/.}, a publicly accessible resource. The dataset contains 200k noisy documents. We only keep the first page of each document, since the invoice information is most likely to show in page one. To reduce the number of noisy samples, we only train on documents if they have 50-300 words (detected by our OCR engine) and more than 3 invoice fields are found by our rule-based method in Sec.~\ref{sec:pseudo_labels}. As a result, we have 129k unlabeled training samples.

For model evaluation, we manually select 350 real invoices as the test set and annotate the 7 fields mentioned above. We note that images of this dataset have lower quality and more clutter background (see Figure~\ref{fig:vis}) which make them more challenging than the IN-Invoice dataset.

More information of the datasets is illustrated in the appendix (Sec.~\ref{sec:appendix}).

\subsection{Evaluation Metric}
We use the macro-average of end-to-end  F1 score over fields as a metric to evaluate models. Specifically, exact string matching between our predicted values and the ground-truth ones is used to count true positive, false positive and false negative. Precision, recall and F1 score is obtained accordingly for each field. The reported scores are averaged over 5 runs to reduce the effect of randomness.

\subsection{Baselines}
It is challenging to compare our method with existing field extraction systems, since they have been evaluated using different datasets in different settings. To the best of our knowledge, there are no existing methods that perform field extraction using only unlabeled data. So, we build the following baselines to validate our method.

\noindent\textbf{Bootstrap Labels (B-Labels)}: the proposed simple rules in Sec.~\ref{sec:pseudo_labels} can be used to do field extraction directly without training data. So, we first show the effectiveness of this method and set up a baseline for later comparison. 

\noindent\textbf{Transformers train with B-Labels}: since we use transformers as the backbone to extract features of words, we train transformer models using the B-Labels as baselines to evaluate the performance gain from (1) the data-driven models in the pipeline and (2) the  refinement module. Both the content of the text and its location are important for field prediction. So, our default transformer backbone is LayoutLM~\cite{xu2020layoutlm} which takes both text and location as input. Further, we also experiment with two popular transformer models, i.e., BERT~\cite{devlin2018bert} and RoBERTa~\cite{liu2019roberta}, which take only text as input.

\subsection{Implementation Details}. Our framework is implemented using Pytorch and the experiments are conducted with Tesla V100 GPUs. We use a commercial OCR engine\footnote{https://api.einstein.ai/signup} to detect words and their locations and use Tesseract\footnote{https://github.com/tesseract-ocr/tesseract} to rank the words in reading order. The key list and data type used in Sec.~\ref{sec:pseudo_labels} for each dataset are shown in Table~\ref{tab:keylist} in Sec.~\ref{sec:appendix}. As we can see, the key lists and data types are quite broad. We set $\alpha$ in Eq.~\ref{eq:keyvaluegeo} to 4.0. To further remove false positives, we remove the value candidates if the localized key is not within its neighboring zone. Specifically, we define the neighboring zone around the value candidate extending all the way to the left of the image, four candidate heights above it and one candidate height below it. We keep the refine branch number $k=3$ for all experiments. We add one hidden FC layer with 768 units before classification when stage number is $>1$. We fix $\beta$ in Eq.~\ref{eq:loss} to be 1.0 for all invoice experiments, except that we use $\beta=5.0$ for BERT-base refinement in Table~\ref{tab:invoice_valid_transformers} due to its better performance in the validation set. For both our model and baselines, we train models for 2 epochs and pick the model with the best F1 score in validation set. To prevent overfitting, we adopt a two-step training strategy, where the pseudo-labels are used to train the first branch of our model and then we fix the first branch along with the feature extractor during the refinement. We set batch size to 8 and use the Adam optimizer with learning rate of 5$e^{-5}$.

\subsection{Comparison with Baselines}
\textbf{Main Comparison}. We primarily validate our design using our IN-Invoice dateset, since it contains large-scale clean, unlabeled training data and sufficient amount of valid/test data. We first validate our method using LayoutLM (our default choice) as the backbone. The comparison results are shown in Table~\ref{tab:invoice_valid} and Table~\ref{tab:invoice_test}. The Bootstrap Labels (B-Labels) baseline achieves 43.8\% and 44.1\% F1 score in valid and test sets, which indicates that our B-Labels have reasonable accuracy, but are still noisy. When we use the B-Labels to train a LayoutLM transformer, we obtain a significant performance improvement, $\sim$15\% increase in valid set and $\sim$17\% in test set. We tried both LayoutLM-base (113M parameters) and LayoutLM-large (343M parameters) models as backbones and we did not see performance improvement when using a larger model. Adding our refinement module significantly improves model precision, $\sim$6\% in valid set and $\sim$7\% in test set, while slightly decreasing the recall, $\sim$2.5\% in valid set and $\sim$3\% in test set. This is because the refine labels  become more and more confident in later stages leading to higher model precision. However, the refinement stage also removes some low confidence false negatives which results in lower recall. Overall, our refinement module further improves performance, resulting in a gain of $\sim3\%$ in F1 score. 

\noindent\textbf{Results with Different Transformers}. We use LayoutLM as the default feature backbone, since both the text and its location is important for our task. To understand the impact of different transformer models as backbone, we experiment with two additional models, BERT and RoBERTa, where only text is used as input. The comparison results are shown in Table~\ref{tab:invoice_valid_transformers} and Table~\ref{tab:invoice_test_transformers}. We have the following observations: (1) we still obtain large improvement when training BERT and RoBERTa directly using our B-Labels and (2) our refinement module consistently improves the baseline results for different transformer choices with different amount of parameters (base or large). Moreover, LayoutLM yields much higher results compared to the other two backbones, which indicates that the text location is indeed very important for obtaining good performance in our task.

\begin{figure*}
    \centering
    \includegraphics[width=1.0\linewidth]{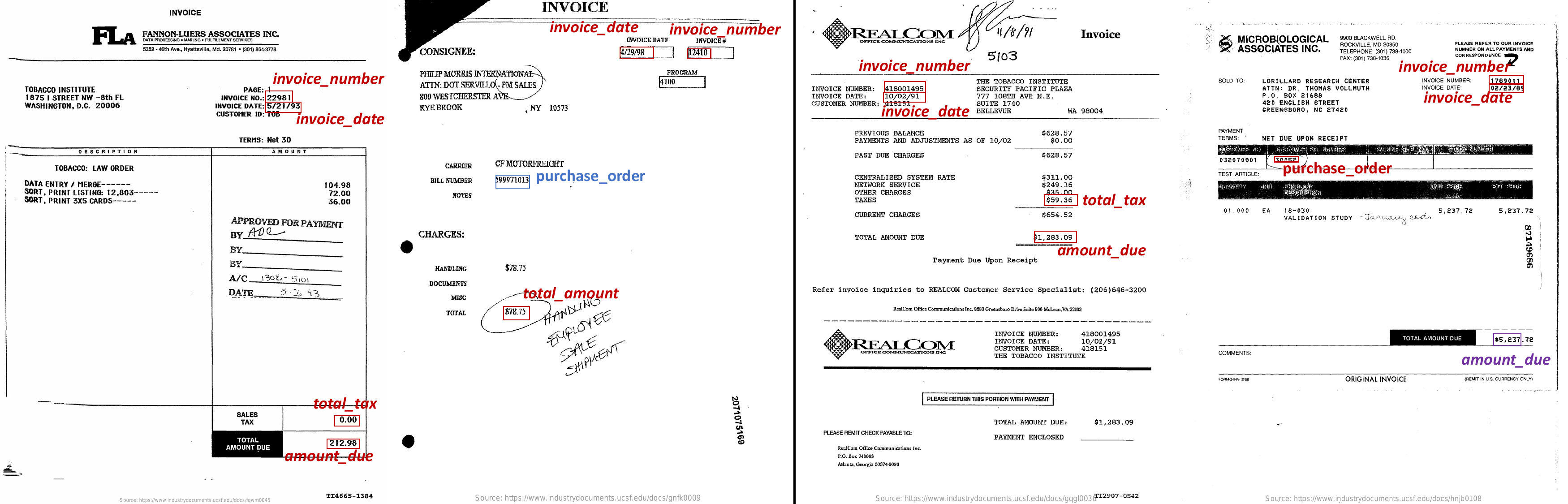}
    \caption{Visualization of our method on the INV-CDIP test set. Correct predictions are marked in red font. Incorrect predictions are marked in blue (due to field extractor error) and purple (due to OCR recognition error).}
    \label{fig:vis}
\end{figure*}

\noindent\textbf{Evaluation on INV-CDIP Test Set}. We evaluate our models trained using IN-Invoice data directly on the introduced INV-CDIP test set in Table~\ref{tab:invoice_public}. Our simple rule-based method obtains 25.1\% F1 score which is reasonable, but much lower compared to the results on our internal IN-Invoice dataset. The reason is that the INV-CDIP test set is visually noisy which results in more OCR recognition errors. The LayoutLM baselines still obtain large improvements over the B-Labels baseline. Also, our refinement module further improves more than 2\% in F1 score. The results suggest that our method adapts well to the new dataset. We show some visualizations in Figure~\ref{fig:vis}. We can see that our method obtains good performance, although the invoices are very diverse across different templates, have cluttered background and are in low resolution. 

\begin{table}[h]
\scalebox{.95}{
\begin{tabular}{l|c|ccc}
\hline
Model & Labels& Prec. & Rec. & F1 \\
\hline
Bootstrap Labels& -- & 40.5 & 50.7 & 43.8 \\
\hline
\hline
LayoutLM-base& \multirow{4}{*}{B-Labels}  &54.8 & 66.6 & 59.2 \\
\textbf{+ PLE} &  &60.9 &64.0 &\textbf{61.9}  \\
\cline{1-1}
\cline{3-5}
LayoutLM-large& & 55.2 &65.6 &58.8 \\
\textbf{+ PLE}& &61.3 &63.2 &\textbf{61.8} \\
\hline
\end{tabular}}
\caption{Comparison with baselines on IN-Invoice valid set. Models are trained using the unlabeled IN-Invoice train set.
}
\label{tab:invoice_valid}
\end{table}

\begin{table}[h]
\scalebox{.95}{
\begin{tabular}{l|c|ccc}
\hline
Model & Labels &Prec. & Rec. & F1 \\
\hline
Bootstrap Labels& -- & 41.0 & 50.9 & 44.1 \\
\hline
\hline
LayoutLM-base& \multirow{4}{*}{B-Labels} & 57.5 &67.6 &61.2  \\
\textbf{+ PLE}&  & 64.7 &64.5 &\textbf{63.8}  \\
\cline{1-1}
\cline{3-5}
LayoutLM-large&  & 58.2 & 67.1 &61.0  \\
\textbf{+ PLE}& & 65.6 &64.0 &\textbf{64.1} \\
\hline
\end{tabular}}
\caption{Comparison with baselines on IN-Invoice test set. Models are trained using the unlabeled IN-Invoice train set.
}
\label{tab:invoice_test}
\end{table}

\begin{table}[h]
\scalebox{.95}{
\begin{tabular}{l|c|cccc}
\hline
Model & Labels& Prec. & Rec. & F1  \\
\hline
Bootstrap Labels& -- & 40.5 & 50.7 & 43.8 \\
\hline
\hline
BERT-base & \multirow{8}{*}{B-Labels}& 48.8 & 59.6 & 52.8 \\
\textbf{+ PLE}& &49.9 & 59.3 & \textbf{53.4}  \\
\cline{1-1}
\cline{3-5}
BERT-large& &53.7 & 60.9 & 56.5  \\
\textbf{+ PLE}&  & 58.0 & 59.4 & \textbf{58.1}  \\
\cline{1-1}
\cline{3-5}
RoBERTa-base& &55.1 & 60.5 & 57.2  \\
\textbf{+ PLE}&  &59.9 & 58.0 & \textbf{58.5}  \\
\cline{1-1}
\cline{3-5}
RoBERTa-large& &55.3 & 61.8 & 57.8  \\
\textbf{+ PLE}&  &60.5 & 60.1 & \textbf{59.1}  \\
\hline
\end{tabular}
}
\caption{Comparison using different transformers on IN-Invoice valid set. Models are trained using the unlabeled IN-Invoice train set.}
\label{tab:invoice_valid_transformers}
\end{table}

\begin{table}[h]
\scalebox{.95}{
\begin{tabular}{l|c|ccc}
\hline
Model & Labels &Prec. & Rec. & F1 \\
\hline
Bootstrap Labels& -- & 41.0 & 50.9 & 44.1 \\
\hline
\hline
BERT-base & \multirow{8}{*}{B-Labels}&51.3 & 61.4 & 54.9  \\
\textbf{+ PLE} &&52.6 & 61.0 & \textbf{55.6}  \\
\cline{1-1}
\cline{3-5}
BERT-large &&55.7 & 61.7 & 57.7  \\
\textbf{+ PLE} & &60.2 & 58.8 & \textbf{58.6}  \\
\cline{1-1}
\cline{3-5}
RoBERTa-base &&57.2 & 61.4 & 58.7  \\
\textbf{+ PLE}  &&62.7 & 58.6 & \textbf{59.8}  \\
\cline{1-1}
\cline{3-5}
RoBERTa-large &&56.3 & 61.4 & 57.8  \\
\textbf{+ PLE} &&62.5 & 59.6 & \textbf{59.2}  \\
\hline
\end{tabular}
}
\caption{Comparison using different transformers on IN-Invoice test set. Models are trained using the unlabeled IN-Invoice train set.}
\label{tab:invoice_test_transformers}
\end{table}

\begin{table}[h]
\scalebox{.95}{
\begin{tabular}{l|c|ccc}
\hline
Model & Labels &Prec. & Rec. & F1 \\
\hline
Bootstrap Labels& -- & 21.8 & 36 & 25.1 \\
\hline
\hline
LayoutLM-base& \multirow{4}{*}{B-Labels} &31.6 & 44.6 & 35.2\\
\textbf{+ PLE}&  & 37.3 & 40.9 & \textbf{37.3}\\
\cline{1-1}
\cline{3-5}
LayoutLM-large&  & 33.8 & 46.5 & 36.9\\
\textbf{+ PLE}& & 40.2 & 42.2 & \textbf{39.4} \\
\hline
\end{tabular}
}
\caption{Comparison with baselines on INV-CDIP test set. Models are trained using the unlabeled IN-Invoice train set.}
\label{tab:invoice_public}
\end{table}
\begin{table}[h]
\scalebox{.95}{
\begin{tabular}{l|c|ccc}
\hline
Model & Eval Set&Prec. & Rec. & F1 \\
\hline
LayoutLM-base& \multirow{2}{*}{\small IN-Inv Val}& 46.3 & 60.6 & 51.4  \\
\textbf{+ PLE} & &51.7 & 58.6 & \textbf{54.2}  \\
\hline
LayoutLM-base& \multirow{2}{*}{\small IN-Inv Test}&47.8 & 62.3 & 52.9 \\
\textbf{+ PLE} & &53.0 & 59.2 & \textbf{55.1} \\
\hline
LayoutLM-base& \multirow{2}{*}{\small INV-CDIP Test}&28.0 & 47.5 & 32.8\\
\textbf{+ PLE} & &31.0 & 46.4 & \textbf{35.4}\\
\hline
\end{tabular}}
\caption{Comparison with baselines when methods are trained using the noisy INV-CDIP train set.}
\label{tab:train_web}
\end{table}

\begin{table}[h]
\begin{tabular}{l|ccc|c}
\hline
Set & R-Labels & 2-step train & B-Labels & F1 \\
\hline
\multirow{4}{*}{Valid} & & \checkmark& \checkmark& 59.7\\
& \checkmark &  & \checkmark& 60.1\\
& \checkmark & \checkmark & & 60.0\\
& \checkmark & \checkmark & \checkmark& \textbf{61.9}\\
\hline
\multirow{4}{*}{Test} & &\checkmark& \checkmark&61.2\\
& \checkmark & & \checkmark& 62.4\\
& \checkmark & \checkmark & & 61.6\\
& \checkmark & \checkmark & \checkmark& \textbf{63.8}\\
\hline
\end{tabular}
\caption{Results of ablation study.}
\label{tab:invoice_ablation}
\end{table}

\noindent\textbf{Learning from Noisy INV-CDIP Data}. Although web data is noisy, it can be freely obtained from the internet. We train our model and the baseline model using the unlabeled train set of the noisy INV-CDIP dataset. The comparison results are shown in Table~\ref{tab:train_web}. As we can see, our method performs well and our PLE module can still improves the baseline by about 2-3\%, although the training set is very noisy.

\subsection{Ablation Study}
\label{sec:ablation}
We conduct ablation study on the IN-Invoice dataset with LayoutLM-base as the backbone.
\begin{figure}[h!]
    \centering
    \includegraphics[width=1.0\linewidth]{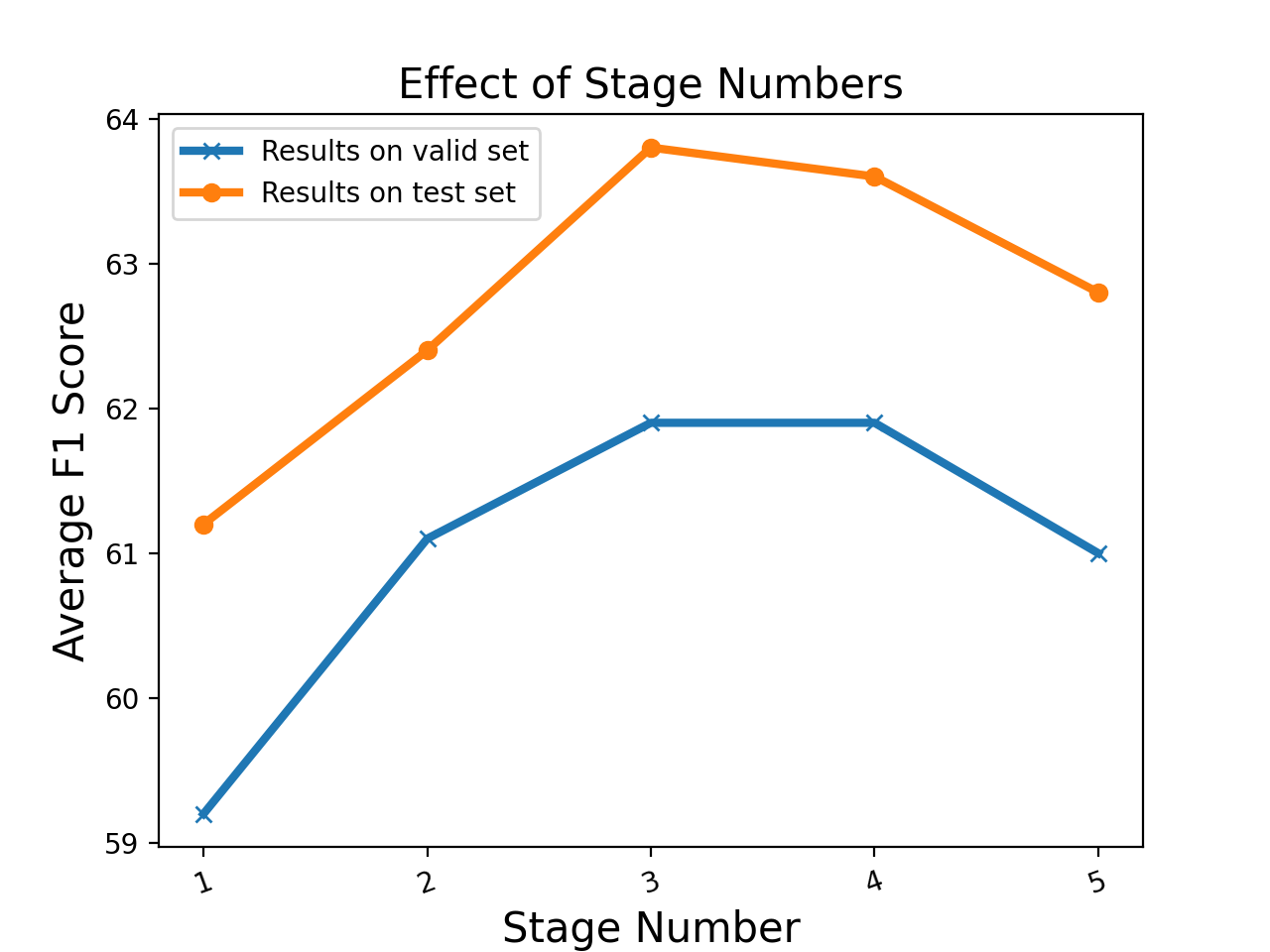}
    \caption{Comparison results with varying stage numbers. When stage number is 1, the model becomes the LayoutLM baseline. }
    \label{fig:stages}
\end{figure}
\begin{figure}[h!]
    \centering
    \includegraphics[width=1.0\linewidth]{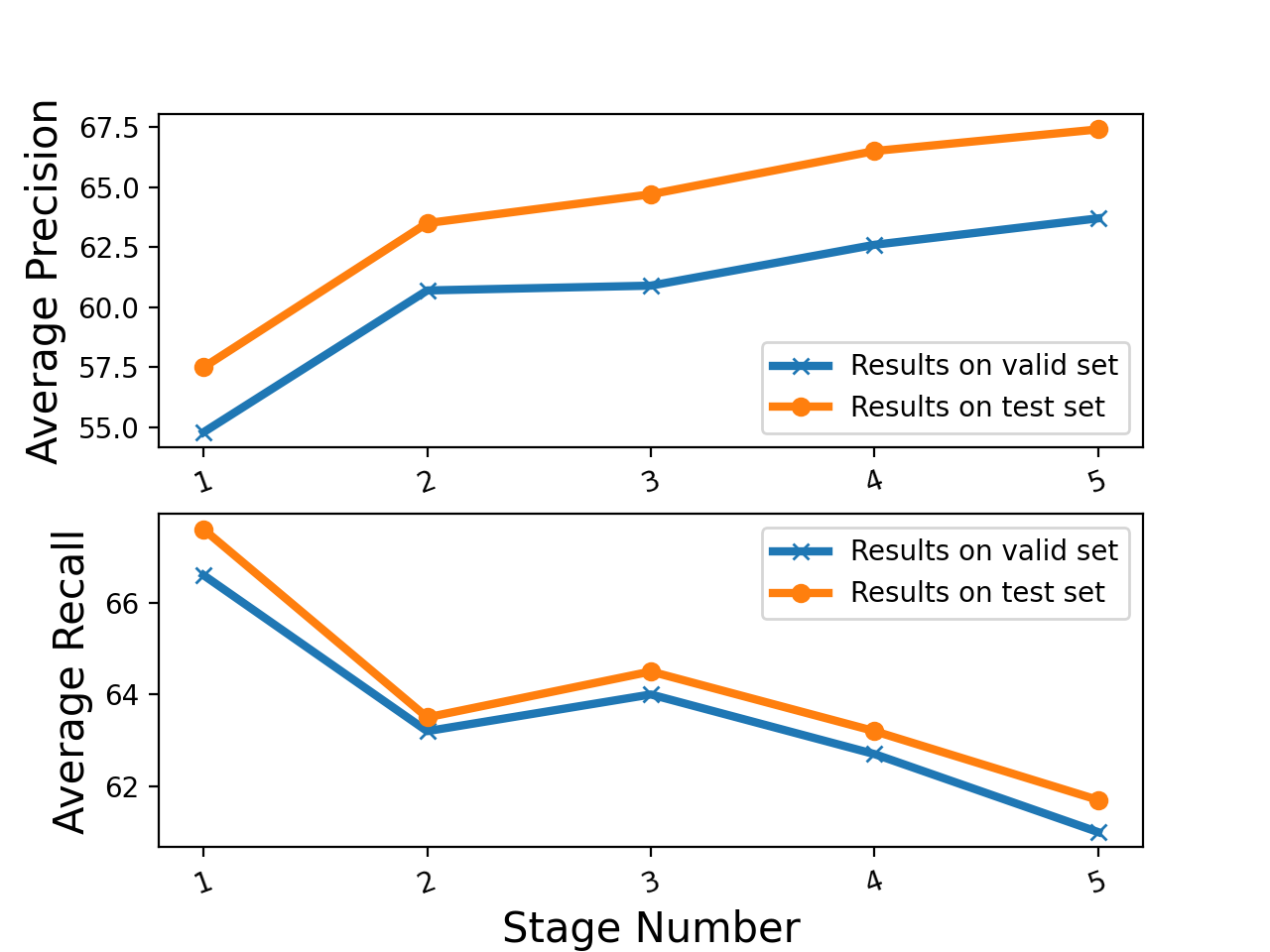}
    \caption{In general, our model has higher precision, but lower recall with larger number of branches.}
    \label{fig:prec_rec_stages}
\end{figure}

\noindent\textbf{Effect of Stage Numbers}. Our model is refined in $k$ stages, where $k=3$ in all experiments. We evaluate our method with varying stage numbers. As we can see in Figure~\ref{fig:stages}, when we increase the stage number, $k$, the model generally performs better on both valid and test sets. The performance with more than one stage is always higher than the single-stage model (our transformer baseline). Model performance reaches the highest when $k=3$. As shown in Figure~\ref{fig:prec_rec_stages}, precision improves while recall drops during model refinement. When $k=3$, we obtain the best balance between precision and recall. When $k>3$ recall drops more than precision improves, leading to a lower F1 score.

\noindent\textbf{Effect of Refined Labels (R-Labels)}. As shown in Figure~\ref{fig:our_method}, the R-Labels obtained in each stage are used in later stages. To analyze the effect of this design, we remove the refined labels in the final loss and only use the B-Labels to train the three branches independently and ensemble the predictions during inference. As shown in Table~\ref{tab:invoice_ablation}, removing refined labels results in 2.2\% and 2.6\% decrease in F1 scores in valid and test sets.

\noindent\textbf{Effect of Regularization with B-Labels}. At each stage, we use B-Labels as a  type of regularization to prevent  the model from  overfitting to the over-confident refined labels. We disable the utilization of B-Labels in the refinement stage by setting $\beta=0$ in Eq.~\ref{eq:loss}. As we can see in Table~\ref{tab:invoice_ablation}, model performance drops $\sim$2\% in F1 score without  this regularization.

\noindent\textbf{Effect of Two-step Training Strategy}. To avoid overfitting to noisy labels, we adopt two-step training strategy, where the transformer backbone with the first branch is trained using B-Labels and then fixed during the refinement. We analyze this effect by training our model in a single step. As shown in Table~\ref{tab:invoice_ablation}, single-step training leads to 1.8\% and 1.4\% F1 score decrease in valid and test sets.

\subsection{Limitations and Future Work}
Our work focuses on training models using unlabeled forms. An interesting future avenue would be to utilize additional labeled data in a semi-supervised setting. Moreover, validating our methods with more form types would  also be valuable. We will consider these topics in future work.

\section{Conclusion}
We proposed a field extraction system that can be trained using forms without field-level annotations. We first introduced a rule-based method to get initial pseudo-labels of forms. Then, we proposed a transformer-based method and improved the model using progressively ensembled labels. We demonstrated that our method outperforms the baselines on invoice datasets and each component of our method makes considerable contribution.

\section{Broader Impact}
This work is potentially useful for improving information extraction systems from forms. So, it has positive impacts including improving document processing efficiency, thus reducing human labor. Reducing human labor may also cause negative consequences such as job loss or displacement, particularly amongst low-skilled labor who may be most in need of gainful employment. The negative impact is not specific to this work and should be addressed broadly in the field of AI research.

We are securely using the IN-Invoice dataset internally. For the public data in the INV-CDIP dataset, we made certain to consider their provenance and there are no restrictions on the use of the public data. All the annotations were conducted and carefully reviewed by the authors. As a result, and given the fact that the datasets contain forms without any personally identifiable data, we have relative confidence that the datasets are ethically sourced.

\bibliography{anthology,custom}

\begin{thebibliography}{22}
\expandafter\ifx\csname natexlab\endcsname\relax\def\natexlab#1{#1}\fi

\bibitem[{Aggarwal et~al.(2020)Aggarwal, Gupta, Sarkar, and
  Krishnamurthy}]{aggarwal2020form2seq}
Milan Aggarwal, Hiresh Gupta, Mausoom Sarkar, and Balaji Krishnamurthy. 2020.
\newblock Form2seq: A framework for higher-order form structure extraction.
\newblock In \emph{EMNLP}.

\bibitem[{Chiticariu et~al.(2013)Chiticariu, Li, and
  Reiss}]{chiticariu-etal-2013-rule}
Laura Chiticariu, Yunyao Li, and Frederick~R. Reiss. 2013.
\newblock Rule-based information extraction is dead! long live rule-based
  information extraction systems!
\newblock In \emph{EMNLP}.

\bibitem[{Denk and Reisswig(2019)}]{denk2019bertgrid}
Timo~I Denk and Christian Reisswig. 2019.
\newblock Bertgrid: Contextualized embedding for 2d document representation and
  understanding.
\newblock \emph{NeurIPS Workshop}.

\bibitem[{Devlin et~al.(2019)Devlin, Chang, Lee, and
  Toutanova}]{devlin2018bert}
Jacob Devlin, Ming-Wei Chang, Kenton Lee, and Kristina Toutanova. 2019.
\newblock Bert: Pre-training of deep bidirectional transformers for language
  understanding.
\newblock \emph{NAACL}.

\bibitem[{Ester et~al.(1996)Ester, Kriegel, Sander, Xu
  et~al.}]{ester1996density}
Martin Ester, Hans-Peter Kriegel, J{\"o}rg Sander, Xiaowei Xu, et~al. 1996.
\newblock A density-based algorithm for discovering clusters in large spatial
  databases with noise.
\newblock In \emph{KDD}.

\bibitem[{Gao et~al.(2021)Gao, Xue, Ramaiah, Xing, Xu, and
  Xiong}]{gao2021value}
Mingfei Gao, Le~Xue, Chetan Ramaiah, Chen Xing, Ran Xu, and Caiming Xiong.
  2021.
\newblock Value retrieval with arbitrary queries for form-like documents.
\newblock \emph{arXiv preprint arXiv:2112.07820}.

\bibitem[{Harley et~al.(2015)Harley, Ufkes, and Derpanis}]{harley2015icdar}
Adam~W Harley, Alex Ufkes, and Konstantinos~G Derpanis. 2015.
\newblock Evaluation of deep convolutional nets for document image
  classification and retrieval.
\newblock In \emph{ICDAR}.

\bibitem[{Huang et~al.(2019)Huang, Chen, He, Bai, Karatzas, Lu, and
  Jawahar}]{huang2019icdar2019}
Zheng Huang, Kai Chen, Jianhua He, Xiang Bai, Dimosthenis Karatzas, Shijian Lu,
  and CV~Jawahar. 2019.
\newblock Icdar2019 competition on scanned receipt ocr and information
  extraction.
\newblock In \emph{ICDAR}.

\bibitem[{Jaume et~al.(2019)Jaume, Ekenel, and Thiran}]{jaume2019funsd}
Guillaume Jaume, Hazim~Kemal Ekenel, and Jean-Philippe Thiran. 2019.
\newblock Funsd: A dataset for form understanding in noisy scanned documents.
\newblock In \emph{ICDARW}.

\bibitem[{Katti et~al.(2018)Katti, Reisswig, Guder, Brarda, Bickel, H{\"o}hne,
  and Faddoul}]{katti2018chargrid}
Anoop~Raveendra Katti, Christian Reisswig, Cordula Guder, Sebastian Brarda,
  Steffen Bickel, Johannes H{\"o}hne, and Jean~Baptiste Faddoul. 2018.
\newblock Chargrid: Towards understanding 2d documents.
\newblock \emph{EMNLP}.

\bibitem[{Liu et~al.(2019)Liu, Ott, Goyal, Du, Joshi, Chen, Levy, Lewis,
  Zettlemoyer, and Stoyanov}]{liu2019roberta}
Yinhan Liu, Myle Ott, Naman Goyal, Jingfei Du, Mandar Joshi, Danqi Chen, Omer
  Levy, Mike Lewis, Luke Zettlemoyer, and Veselin Stoyanov. 2019.
\newblock Roberta: A robustly optimized bert pretraining approach.
\newblock \emph{arXiv preprint arXiv:1907.11692}.

\bibitem[{Majumder et~al.(2020)Majumder, Potti, Tata, Wendt, Zhao, and
  Najork}]{majumder2020representation}
Bodhisattwa~Prasad Majumder, Navneet Potti, Sandeep Tata, James~Bradley Wendt,
  Qi~Zhao, and Marc Najork. 2020.
\newblock Representation learning for information extraction from form-like
  documents.
\newblock In \emph{ACL}.

\bibitem[{Mathew et~al.(2021)Mathew, Karatzas, and Jawahar}]{mathew2021docvqa}
Minesh Mathew, Dimosthenis Karatzas, and CV~Jawahar. 2021.
\newblock Docvqa: A dataset for vqa on document images.
\newblock In \emph{WACV}.

\bibitem[{Palm et~al.(2019)Palm, Laws, and Winther}]{palm2019attend}
Rasmus~Berg Palm, Florian Laws, and Ole Winther. 2019.
\newblock Attend, copy, parse end-to-end information extraction from documents.
\newblock In \emph{ICDAR}.

\bibitem[{Park et~al.(2019)Park, Shin, Lee, Lee, Surh, Seo, and
  Lee}]{park2019cord}
Seunghyun Park, Seung Shin, Bado Lee, Junyeop Lee, Jaeheung Surh, Minjoon Seo,
  and Hwalsuk Lee. 2019.
\newblock Cord: A consolidated receipt dataset for post-ocr parsing.

\bibitem[{Schuster et~al.(2013)Schuster, Muthmann, Esser, Schill, Berger,
  Weidling, Aliyev, and Hofmeier}]{6628593}
Daniel Schuster, Klemens Muthmann, Daniel Esser, Alexander Schill, Michael
  Berger, Christoph Weidling, Kamil Aliyev, and Andreas Hofmeier. 2013.
\newblock Intellix -- end-user trained information extraction for document
  archiving.
\newblock In \emph{ICDAR}.

\bibitem[{Tang et~al.(2017)Tang, Wang, Bai, and Liu}]{tang2017multiple}
Peng Tang, Xinggang Wang, Xiang Bai, and Wenyu Liu. 2017.
\newblock Multiple instance detection network with online instance classifier
  refinement.
\newblock In \emph{CVPR}.

\bibitem[{Vaswani et~al.(2017)Vaswani, Shazeer, Parmar, Uszkoreit, Jones,
  Gomez, Kaiser, and Polosukhin}]{vaswani2017attention}
Ashish Vaswani, Noam Shazeer, Niki Parmar, Jakob Uszkoreit, Llion Jones,
  Aidan~N Gomez, Lukasz Kaiser, and Illia Polosukhin. 2017.
\newblock Attention is all you need.
\newblock \emph{NeurIPS}.

\bibitem[{Wang et~al.(2020)Wang, Zhan, Liu, and Liang}]{wang2020docstruct}
Zilong Wang, Mingjie Zhan, Xuebo Liu, and Ding Liang. 2020.
\newblock Docstruct: A multimodal method to extract hierarchy structure in
  document for general form understanding.
\newblock \emph{EMNLP}.

\bibitem[{Winkler(1990)}]{winkler1990string}
William~E Winkler. 1990.
\newblock String comparator metrics and enhanced decision rules in the
  fellegi-sunter model of record linkage.

\bibitem[{Xu et~al.(2020)Xu, Li, Cui, Huang, Wei, and Zhou}]{xu2020layoutlm}
Yiheng Xu, Minghao Li, Lei Cui, Shaohan Huang, Furu Wei, and Ming Zhou. 2020.
\newblock Layoutlm: Pre-training of text and layout for document image
  understanding.
\newblock In \emph{KDD}.

\bibitem[{Xue et~al.(2021)Xue, Gao, Chen, Xiong, and Xu}]{xue2021robustness}
Le~Xue, Mingfei Gao, Zeyuan Chen, Caiming Xiong, and Ran Xu. 2021.
\newblock Robustness evaluation of transformer-based form field extractors via
  form attacks.
\newblock \emph{arXiv preprint arXiv:2110.04413}.

\end{thebibliography}
\bibliographystyle{acl_natbib}

\appendix

\section{Appendix}
\label{sec:appendix}

\renewcommand{\thefigure}{A\arabic{figure}}
\setcounter{figure}{0}

\setcounter{table}{0}
\renewcommand{\thetable}{A\arabic{table}}

\begin{figure}[h!]
    \centering
    \includegraphics[width=1.0\linewidth]{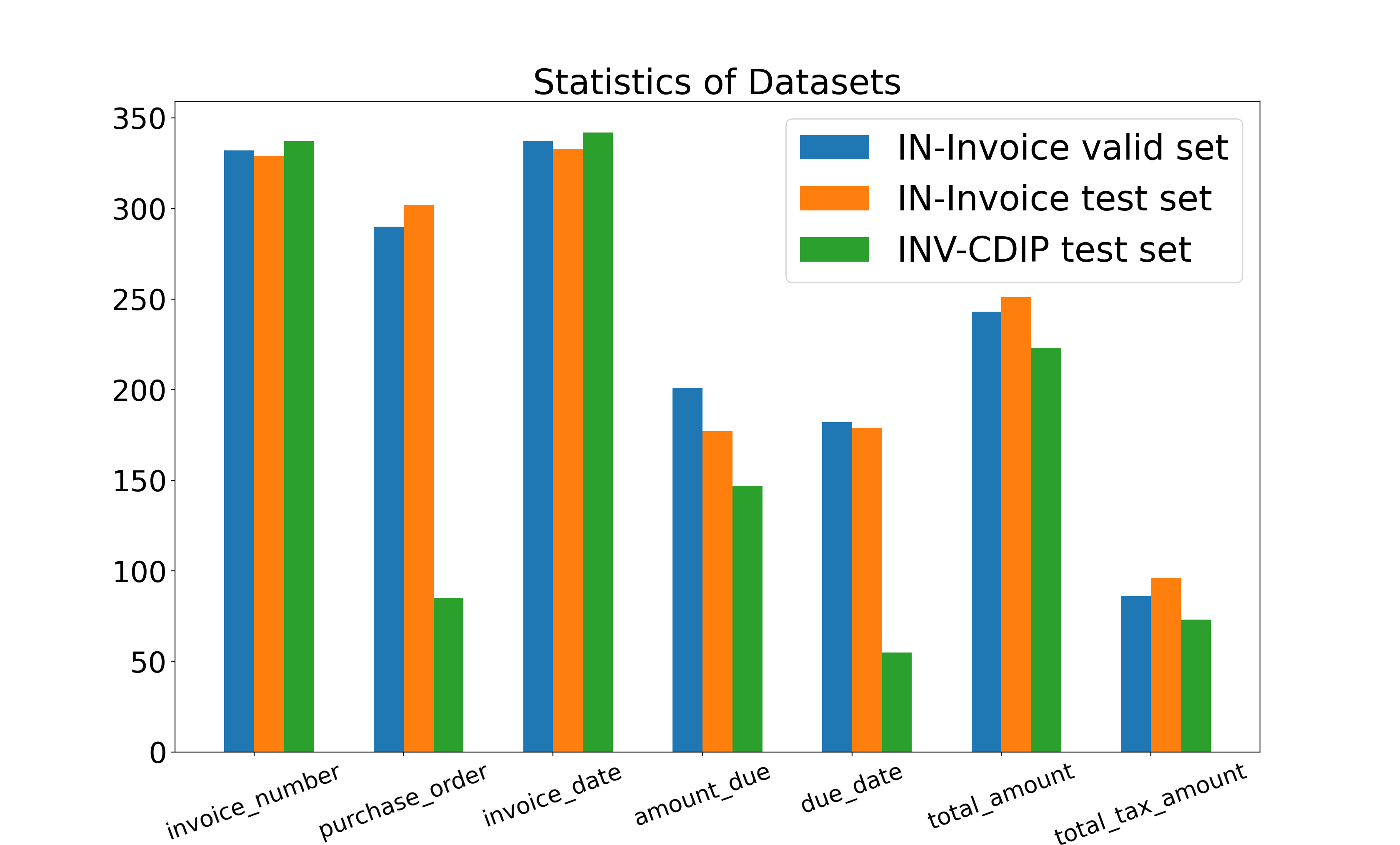}
    \caption{Field statistics of the datasets. 
    }
    \label{fig:dataset_stats}
\end{figure}

\begin{table*}[t]
    \centering
    \begin{tabular}{l|c|c}
    \hline
    Field & Data Type & Key List \\
    \hline
    inv$\_$number & number &\small ["invoice number", "invoice $\#$", "invoice", "invoice no.", "invoice no"] \\
    po$\_$number & number & \small["po $\#$", "po number", "p.o. $\#$", "p.o. number", "po", "purchase order number"] \\
    inv$\_$date & date & \small["date", "invoice date:", "invoice date"] \\
    due$\_$date & date & \small["due date"] \\
    total$\_$amount & number/money& \small["total", "invoice total"] \\
    due$\_$amount & number/money & \small["amount due", "balance due"] \\
    total$\_$tax & number/money & \small["tax"] \\
    \hline
    \end{tabular}
    \caption{Key list and data type used in our experiments.}
    \label{tab:keylist}
\end{table*}
\begin{figure*}[htp]
    \centering
    \includegraphics[width=1.0\linewidth]{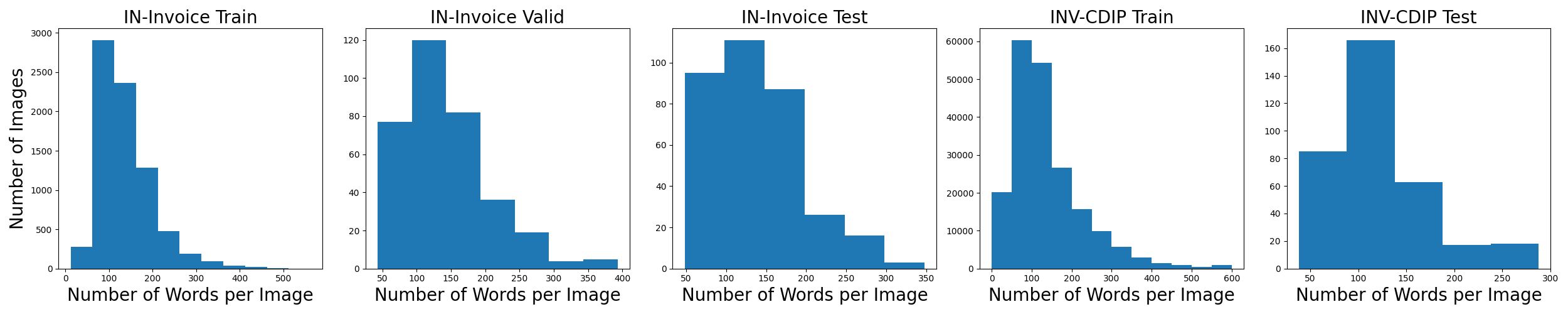}
    \caption{Number of words per image of the datasets. 
    }
    \label{fig:word_cnt}
\end{figure*}
\begin{figure*}[htp]
    \centering
    \includegraphics[width=1.0\linewidth]{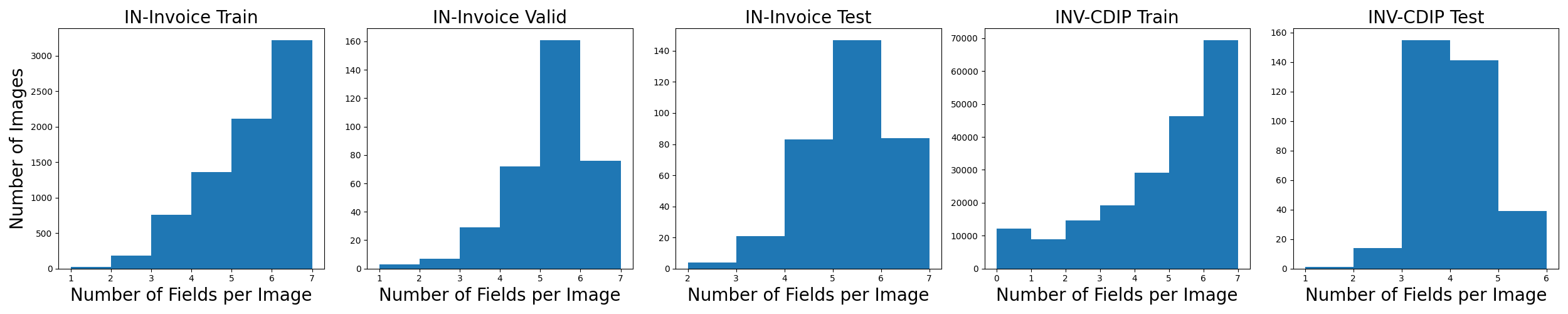}
    \caption{Number of fields per image of the datasets. 
    }
    \label{fig:field_cnt}
\end{figure*}

\subsection{More Information about Datasets}
The field statistics of both IN-Invoice and INV-CDIP datasets are shown in Figure~\ref{fig:dataset_stats}. As we can see, the validation and test sets of our internal IN-Invoice dataset have a similar statistical distribution of fields, while the public INV-CDIP test set is different. 

Moreover, the number of words per image in each dataset is shown in Figure~\ref{fig:word_cnt}. As shown, most images have 50-300 words and images with 100-150 words are typical in all the sets. We compute the number of fields per image in Figure~\ref{fig:field_cnt}. As we can see, there are averagely more fields annotated per image in IN-Invoice valid/test sets than those annotated in the INV-CDIP test set. Note that there are no field annotations in the train sets. We show the matched pseudo-labels in the train sets in Figure~\ref{fig:field_cnt}. As we can see, the number of matched pseudo-labels per image in the IN-Invoice train set is similar to that in the INV-CDIP train set. 

Note that all data described was collected exclusively for the academic purpose of conducting research. The purpose of using the invoices and data was only for training and evaluating the model. No data is stored upon completion of the research process. 
\end{document}